\documentclass[sigconf]{acmart}
\AtBeginDocument{%
  \providecommand\BibTeX{{%
    \normalfont B\kern-0.5em{\scshape i\kern-0.25em b}\kern-0.8em\TeX}}}

\setcopyright{acmcopyright}
\copyrightyear{2023}
\acmYear{2023}
\acmDOI{XXXXXXX.XXXXXXX}

\acmConference[ASP-DAC'23]{ASP-DAC'23: 28th Asia and South Pacific Design Automation Conference}{Jan. 16-19, 2023}{Tokyo, Japan}
\acmBooktitle{ASP-DAC'23: 28th Asia and South Pacific Design Automation Conference, Jan. 16-19, 2023, Tokyo, Japan} 
\acmPrice{15.00}
\acmISBN{978-1-4503-XXXX-X/18/06}

\usepackage{hyperref}
\usepackage{multirow}
\newenvironment{myitemize}{\begin{list}{$\bullet$}
{\setlength{\topsep}{1mm}
\setlength{\itemsep}{0.25mm}
\setlength{\parsep}{0.25mm}
\setlength{\itemindent}{0mm}
\setlength{\partopsep}{0mm}
\setlength{\labelwidth}{15mm}
\setlength{\leftmargin}{4mm}}}{\end{list}}

\usepackage{array}
\newcommand{\PreserveBackslash}[1]{\let\temp=\\#1\let\\=\temp}
\newcolumntype{C}[1]{>{\PreserveBackslash\centering}p{#1}}
\newcolumntype{R}[1]{>{\PreserveBackslash\raggedleft}p{#1}}
\newcolumntype{L}[1]{>{\PreserveBackslash\raggedright}p{#1}}

\renewcommand\footnotetextcopyrightpermission[1]{}

\begin{document}

\title{Safety-driven Interactive Planning for Neural Network-based Lane Changing}

\author{Xiangguo Liu}
\orcid{0000-0002-1944-3923}
\affiliation{%
  \institution{Northwestern University}
  \city{Evanston}
  \state{IL}
  \country{USA}
  \postcode{60201}
}
\email{xg.liu@u.northwestern.edu}

\author{Ruochen Jiao}
\affiliation{%
  \institution{Northwestern University}
  \city{Evanston}
  \state{IL}
  \country{USA}}
\email{RuochenJiao2024@u.northwestern.edu}

\author{Bowen Zheng}
\affiliation{%
 \institution{Pony.ai}
 \city{Fremont}
 \state{CA}
 \country{USA}}
\email{bowen.zheng@pony.ai}

\author{Dave Liang}
\affiliation{%
 \institution{Pony.ai}
 \city{Fremont}
 \state{CA}
 \country{USA}}
\email{dave.liang@pony.ai}

\author{Qi Zhu}
\affiliation{%
  \institution{Northwestern University}
  \city{Evanston}
  \state{IL}
  \country{USA}
}
\email{qzhu@northwestern.edu}

\renewcommand{\shortauthors}{Liu, et al.}

\begin{abstract}
Neural network-based driving planners have shown great promises in improving task performance of autonomous driving. However, it is critical and yet very challenging to ensure the safety of systems with neural network-based components, especially in dense and highly interactive traffic environments. In this work, we propose a safety-driven interactive planning framework for neural network-based lane changing. To prevent over-conservative planning, we identify the driving behavior of surrounding vehicles and assess their aggressiveness, and then adapt the planned trajectory for the ego vehicle accordingly in an interactive manner. The ego vehicle can proceed to change lanes if a safe evasion trajectory exists even in the predicted worst case; otherwise, it can stay around the current lateral position or return back to the original lane. We quantitatively demonstrate the effectiveness of our planner design and its advantage over baseline methods through extensive simulations with diverse and comprehensive experimental settings, as well as in real-world scenarios collected by an autonomous vehicle company.
\end{abstract}





\maketitle

\section{Introduction}


Lane changing in dense traffic is a very challenging task in autonomous driving, especially in scenarios with complex inter-vehicle interactions. It is a common safety-efficiency dilemma. Some planners set large buffer space for safety~\cite{cheng2021general} to handle uncertainties from surrounding vehicles and the environment, however could be overly conservative and inefficient. Other planners put more emphasis on efficiency and task success rate, but risk safety. It could be even more challenging during the transition period to a fully-automated transportation system, when human-driven and autonomous vehicles need to share the transportation network and interact with each other~\cite{liu2020impact}. Without an accurate estimation of other vehicles' intention, the lane changing process could be either inefficient or unsafe.

Moreover, neural network-based machine learning techniques~\cite{zhu2021safety,luo2021credibility,jiao2022tae,jiao2022semi} have been increasingly utilized in autonomous driving for perception, prediction, planning, etc. Compared with traditional rule-based planners, neural network-based ones have the potential to significantly improve performance and efficiently handle complex scenarios~\cite{cao2020reinforcement} such as those in mandatory lane changing. However, the learning components also increase the difficulty in ensuring system safety. In the literature, a number of neural network-based methods~\cite{cao2020reinforcement,naveed2020trajectory} measure system performance and safety via extensive simulations, but do not provide safety assurance. On the other hand, while state-of-the-art safety verification techniques~\cite{tran2019safety,huang2019reachnn,ivanov2019verisig} can theoretically analyze learning-enabled systems, they are challenging to scale to complex scenarios and often result in conservative conclusions.

To overcome these challenges, we propose a \emph{safety-driven interactive planning framework with neural network-based planners} in dense lane changing scenarios. Specifically, we have two neural network planners for longitudinal and lateral motions, respectively. The two planners take motion status of surrounding vehicles and the ego vehicle as input, and output planned accelerations for the ego vehicle. In order to enhance safety while improve efficiency, the ego vehicle can make lane changing attempt under the neural network planners only if it has a safe evasion trajectory even in the predicted worst case; and if such safe evasion trajectory does not exist, the planned trajectory from neural networks will be adjusted according to safety analysis of all involved vehicles. To prevent an overly conservative planner design, we leverage another neural network to assess the aggressiveness of the following vehicle in the target lane and predict whether it is willing to let the ego vehicle complete the lane change. In the case that the following vehicle is cautious (intuitively meaning that it is willing to let the ego vehicle get in front of it), the ego vehicle can complete the lane changing confidently; otherwise, the following vehicle is aggressive and the ego vehicle needs to be more conservative.

\begin{figure}[tbp]
\centering
\includegraphics[scale=0.45]{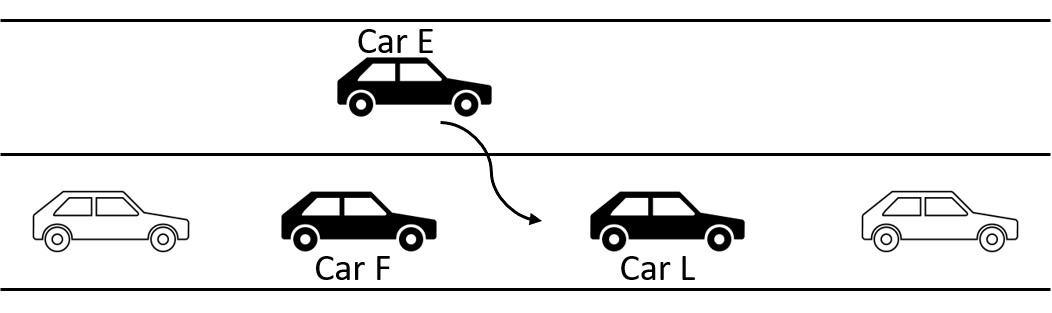}
\caption{Lane changing scenario. The ego vehicle $E$, an autonomous vehicle, intends to change lanes and insert itself downstream of vehicle $F$, a human-driven or an autonomous vehicle in the target lane.}
\label{fig:scenario}
\end{figure}

Fig.~\ref{fig:scenario} shows the specific scenario we consider. The ego vehicle $E$, an autonomous vehicle, intends to change lanes and insert itself downstream of vehicle $F$, a human-driven or an autonomous vehicle in the target lane. We assume that the worst case occurs when the leading vehicle $L$ in the target lane decelerates abruptly and at the same time the following vehicle $F$ has the highest acceleration under its predicted cautious/aggressive mode.  
A safe evasion trajectory exists if the ego vehicle can take that trajectory and return to the original lane without colliding with other vehicles. 

Specifically, the contributions of our work include:
\begin{myitemize}
\item We propose a safety-driven interactive lane changing framework with neural network-based planners. The framework includes a safety-driven behavior adjustment module that takes the outputs from two neural network-based planners and decides whether to proceed, abort, or hesitate. This is based on analyzing whether a safe evasion trajectory exists, considering the aggressiveness of the following vehicle in the target lane. 
\item Our framework includes a learning-based module for assessing the aggressiveness of the following vehicle in the target lane to prevent over-conservative planning, which improves lane changing success rate and efficiency, especially in dense traffic. 
\item We demonstrate the advantages of our framework on various metrics through extensive simulations on synthetic examples and real-world scenarios, when compared with a traditional optimization-based planner and an end-to-end neural network planner. In particular, our framework is safe in all simulations. Moreover, our framework is \emph{guaranteed} to be safe \emph{if} the aggressive assessment is accurate or if we choose to always treat the following vehicle as aggressive. 
\end{myitemize}


The rest of this paper is organized as follows. In Section~\ref{sec:related_work}, we review related works on lane changing, inter-vehicle interaction, and neural network-based planning. In Section~\ref{sec:Methodology}, we present our safety-driven interactive planning framework. Section~\ref{sec:experiments} shows the experimental results and Section~\ref{sec:conclusion} concludes the paper.




\section{Related Work}\label{sec:related_work}

There is a rich literature for trajectory planning, and specifically, lane changing, which is reviewed in details in~\cite{zheng2014recent,liu2022markov,liu2021trajectory}. For instance, \cite{luo2016dynamic} proposes a dynamic lane changing planner that updates its reference trajectory periodically. If necessary, it can plan a trajectory back to the original lane to eliminate collision. 
However, \cite{luo2016dynamic} assume that the leading and following vehicles in the target lane will keep their velocities unchanged when the ego vehicle changes lanes, which may not hold due to the fluctuation of traffic stream and the interactions between vehicles. 



To improve efficiency and lane changing success rate, especially in dense traffic, prior works~\cite{burger2020interaction,isele2019interactive} have emphasized the importance of modelling inter-vehicle interactions. 
\cite{awal2015efficient} assumes that all vehicles are connected and cooperative, which is not the case during the transition period. \cite{bouton2019cooperation} leverages partially observable Markov decision process to model the level of cooperation of other drivers, and incorporates this belief into reinforcement learning based planner for higher merging success rate. 
In this work, we design a neural network to predict the acceleration of the following vehicle in the target lane. By comparing our prediction with its true acceleration, we can assess the following vehicle's aggressiveness. In the case where there is low confidence with the prediction, the ego vehicle conservatively assumes that the following vehicle is aggressive for making safe decisions.


Machine learning techniques, particularly those based on neural networks, have been increasingly applied to the lane changing task. \cite{ye2020automated} leverages reinforcement learning and considers the possible action of aborting lane changing and returning back to the original lane. 
\cite{cao2020reinforcement} proposes a hierarchical reinforcement and imitation learning (H-REIL) approach that consists of low-level policies learned by imitation learning under different driving modes and a high-level policy learned by reinforcement learning for switching between driving modes. However, these methods do not provide safety assurance, while our method can if our aggressive assessment is accurate or if we choose to always treat the following vehicle in the target lane as aggressive.

There are several works that try to provide formal safety guarantees for lane changing. For instance, \cite{naumann2019provably} analyzes the distance between vehicles to ensure safety, however the distance is derived only based on braking behavior. 
Moreover, it does not explicitly analyze the intention of the following vehicle in the target lane. 
The work in~\cite{chandru2017safe} analyzes the minimum critical distance around surrounding vehicles by considering both braking and steering behavior, and assumes that the worst case occurs when the leading vehicle in the target lane has a full stop suddenly or the following vehicle in the target lane remains its acceleration to close the gap. However, different from our work, \cite{chandru2017safe} neglects the fact that the ego vehicle can steer and brake at the same time to avoid collision, and that the worst case for the leading and the following vehicles can occur at the same time. Moreover, it does not consider inter-vehicle interactions as our approach.

\section{Our Safety-Driven Interactive Planning Framework}\label{sec:Methodology}

\begin{figure}[hbp]
\centering
\includegraphics[scale=0.42]{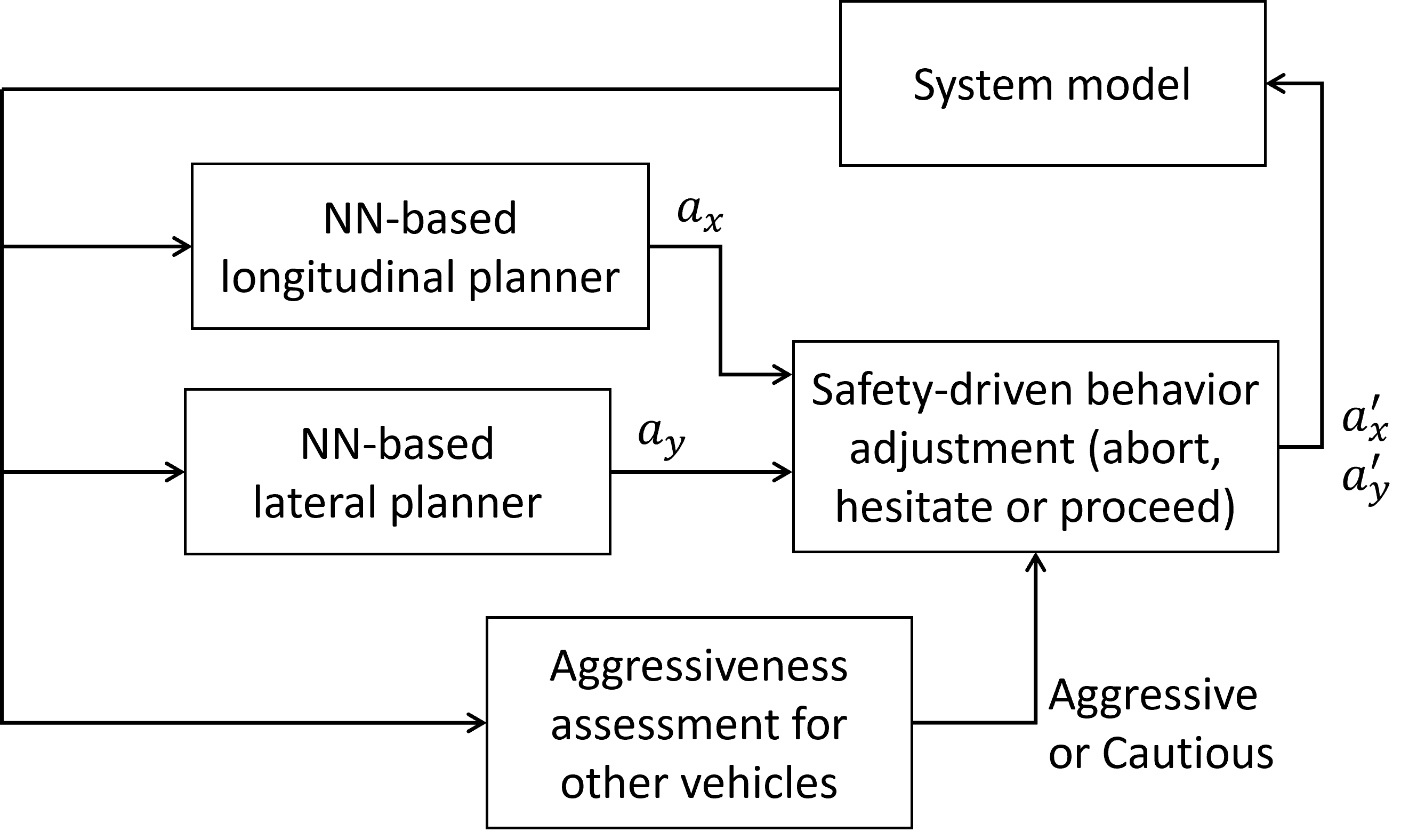}
\caption{Design of our safety-driven interactive planning framework for neural network-based lane changing. Safety-driven behavior adjustment module will adjust risky motions based on analyzing whether a safe evasion trajectory exists, considering the aggressiveness assessment of other vehicles. 
}
\label{fig:framework}
\end{figure}

Our proposed framework design is shown in Fig.~\ref{fig:framework}. In this framework, we consider neural network-based planners for longitudinal and lateral motion planning, with more details in Section~\ref{sec:nn_planners}. To improve lane changing success rate in dense traffic, we leverage another neural network to assess the aggressiveness of the following vehicle $F$ in the target lane, which is discussed in Section~\ref{sec:agg}. Then, based on the predicted behavior of vehicle $F$ (aggressive or cautious), we conduct safety analysis and compute the critical region to avoid collision in the predicted worst case. The trajectory planned under neural networks is adjusted in advance if there is any possibility of collision during the lane changing process. This is detailed in Section~\ref{sec:adj}. 

\subsection{Longitudinal and Lateral Planners}\label{sec:nn_planners}

The two neural networks for longitudinal and lateral planning each has seven independent input variables\footnote{There are eight input variables in Table~\ref{table: nn variables}. However, among $p_x$, $p_{x,l}$ and $p_{x,f}$, only two of them are independent.} and one of the two output variables, as summarized in Table~\ref{table: nn variables}. In order to cover as many traffic scenarios as possible, we synthesize the dataset by simulations according to human driving norm~\cite{liu2022physics}. 
The ego vehicle initially is at the center of the original lane and has no lateral speed. 
Then it changes lanes under an MPC controller~\cite{2020arXiv200108620L} with step size $\delta t=0.1$ second. The optimization goal is to minimize fuel consumption and lane changing time while satisfying safety and comfort constraints.

The dataset collects system states and accelerations of the ego vehicle at every step, which is composed of about 36 million entries. The neural network planner is trained to minimize mean squared error with the Adam optimizer. Note that using more comprehensive datasets with good performance (either from human driving trajectories or from synthesized trajectories), the neural network-based planners can be further improved.




\begin{table}[]
\centering
\caption{Input and output of neural network-based planners.}
\label{table: nn variables}
\begin{tabular}{cc}
\hline
\multicolumn{1}{c|}{notation} & definition \\ \hline
\multicolumn{1}{c}{inputs}              \\ \hline
\multicolumn{1}{c|}{$p_{x}$}         &  longitudinal position of the ego vehicle         \\
\multicolumn{1}{c|}{$p_{y}$}         &    lateral position of the ego vehicle        \\
\multicolumn{1}{c|}{$v_{x}$}         &    longitudinal velocity of the ego vehicle        \\
\multicolumn{1}{c|}{$v_{y}$}         &    lateral velocity of the ego vehicle        \\
\multicolumn{1}{c|}{$p_{x,l}$}         &  longitudinal position of the leading vehicle $L$          \\
\multicolumn{1}{c|}{$v_{x,l}$}         &  longitudinal velocity of the leading vehicle $L$          \\
\multicolumn{1}{c|}{$p_{x,f}$}         &  longitudinal position of the following vehicle $L$          \\
\multicolumn{1}{c|}{$v_{x,f}$}         &  longitudinal velocity of the following vehicle $L$          \\ \hline
\multicolumn{1}{c}{outputs}              \\ \hline
\multicolumn{1}{c|}{$a_{x}$}         &    longitudinal acceleration of the ego vehicle        \\
\multicolumn{1}{c|}{$a_{y}$}         &    lateral acceleration of the ego vehicle        \\ \hline
\end{tabular}
\end{table}


\subsection{Aggressiveness Assessment and Behavior Prediction}\label{sec:agg}

According to~\cite{li2020interaction}, the driving behavior of the following vehicle in the target lane follows one model when it is cautious and another model when it is aggressive. This assumption is validated by real-world human driving data. In this work, we use similar assumptions. We assume that the following vehicle $F$ follows the ego vehicle $E$ when it is cautious and follows the leading vehicle $L$ when it is aggressive. 

For both two cases, we leverage a neural network to predict the accelerations of the following vehicle. Let $a_1$ and $a_0$ denote the accelerations when it is cautious or aggressive, respectively. Different from the motion planner, we do not need $p_y$ and $v_y$ as input variables of the neural network. 

For this prediction task, we also synthesize the dataset via simulations. We generate various types of traffic states for the three vehicles and compute the accelerations of the following vehicle $F$ with the Intelligent Driver Model (IDM)~\cite{jin2016optimal}. The parameters in the IDM model are uniformly sampled by following $a_{x,a}=4 \text{ m}/\text{s}^2$, $v_m=\dot h + v_{x,f}$, $5\text{ m} \leq h_s \leq 8\text{ m}$, $1\text{ s} \leq t_{g} \leq 2\text{ s}$, $a_{x,d}=6 \text{ m}/\text{s}^2$. With the dataset of one million entries, we train the neural network to minimize mean squared error with the Adam optimizer.

The following vehicle's behavior is predicted by comparing its true acceleration $a_{x,f}^*$ with the predicted $a_1$ and $a_0$. When $a_{x,f}^*$ is closer to $a_1$, the following vehicle $F$ is predicted as cautious and follows the ego vehicle $E$; when $a_{x,f}^*$ is closer to $a_0$, it is predicted as aggressive and follows the leading vehicle $L$:
\begin{equation}\label{eq:behavior}
\begin{cases}
|a_{x,f}^*-a_1|<|a_{x,f}^*-a_0|-a_{th}\rightarrow \text{vehicle $F$ is cautious}
\\
|a_{x,f}^*-a_0|<|a_{x,f}^*-a_1|-a_{th}\rightarrow \text{vehicle $F$ is aggressive}
\\
-a_{th} \leq |a_{x,f}^*-a_0|-|a_{x,f}^*-a_1| \leq a_{th}\rightarrow \text{uncertain}
\end{cases}
\end{equation}
Here $a_{th}$ is a threshold. Larger $a_{th}$ means higher confidence on the behavior prediction. For those uncertain scenarios, we assume that the vehicle $F$ is aggressive so that the planned trajectory for ego vehicle is conservative and safe.

Based on the predicted behavior of the following vehicle, we conduct safety analysis. We assume that (1) if the following vehicle is cautious and willing to create gap for ego vehicle, it can at least decelerate with $a_{x,f,d}=6 \text{ m}/\text{s}^2$; and (2) if the following vehicle is aggressive, in the worst case, it can accelerate with $a_{x,f,a}=4 \text{ m}/\text{s}^2$ to prevent ego vehicle from cutting in.






\subsection{Safety Analysis and Motion Adjustment}\label{sec:adj}

\begin{figure}[tbp]
\centering
\includegraphics[scale=0.50]{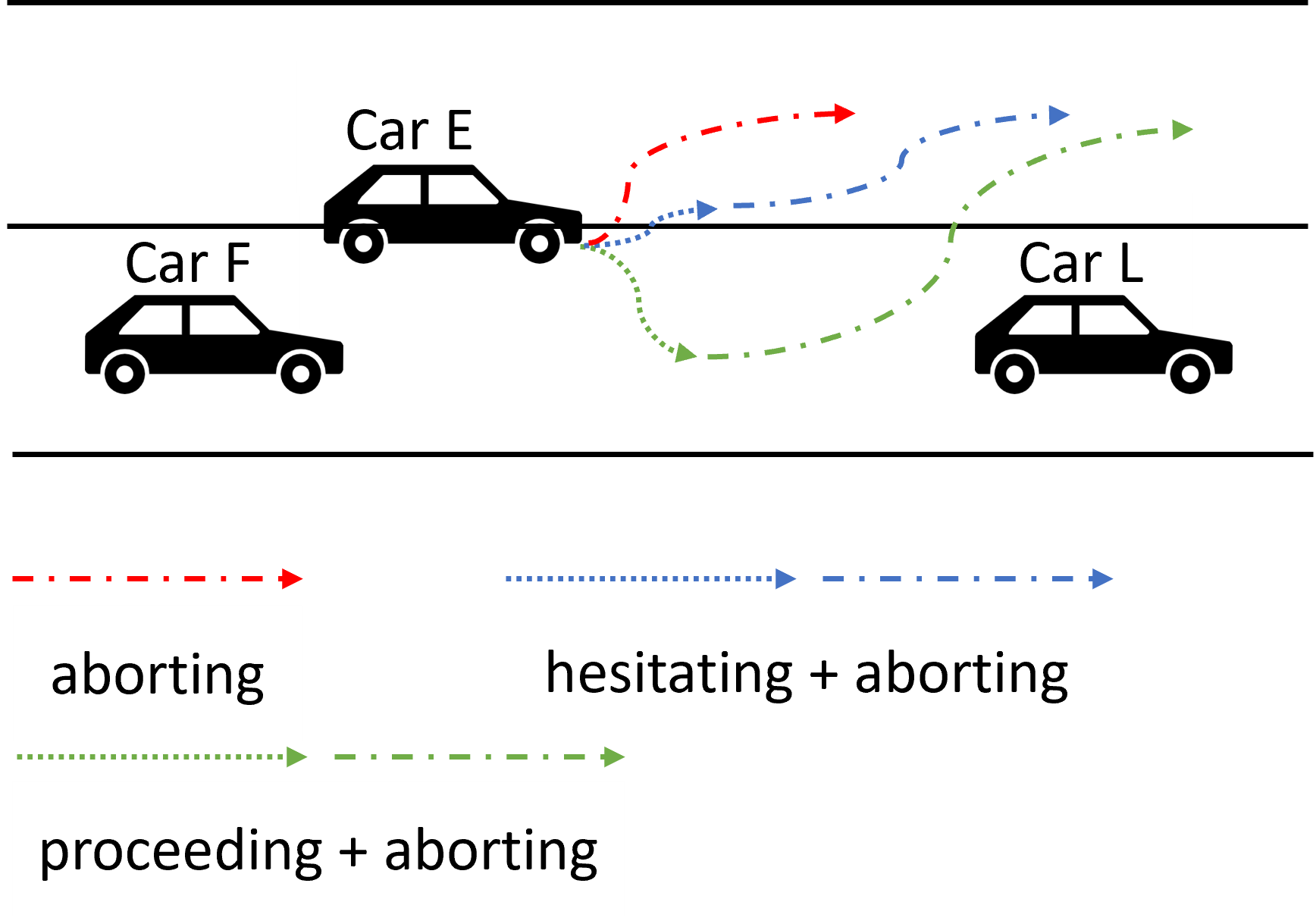}
\caption{At every step in the lane changing process, the ego vehicle has three strategy choices. The first is proceeding to change lanes, in which case the short-term proceeding trajectory (green dotted line) and following complete aborting trajectory (green dash-dotted line) should be verified with safety guarantee. If the first strategy is not safe, the ego vehicle can hesitate around the current lateral position, and we need to verify safety for the short-term hesitating trajectory (blue dotted line) and following complete aborting trajectory (blue dash-dotted line). If both strategies do not work, the ego vehicle can directly abort lane changing behavior and go back to the original lane (red dash-dotted line), which is already verified to be safe in the last planning step.}
\label{fig:strategy}
\end{figure}

We assume that the ego vehicle is safe when it is in the original lane at the very beginning. At every step during the lane changing, the ego vehicle has a safe evasion trajectory computed at the last step. As shown in Fig.~\ref{fig:strategy}, it has three options with decreasing preference: proceed to change lanes, hesitate around the current lateral position, or abort the lane changing and return back to the original lane. It analyzes the state after executing selected behavior for one time step. If it has a safe evasion trajectory following that, it can go ahead with the selected behavior; otherwise, it has to attempt a less preferred behavior. In summary, safety of the ego vehicle is ensured by only selecting the strategy with a following safe evasion trajectory, assuming that the aggressiveness assessment for the following vehicle is correct (if not, safety can only be ensured if the following vehicle is always treated as aggressive).

The behavior of proceeding to change lanes is to directly follow the longitudinal and lateral accelerations computed using the neural network planners. For hesitating, the lateral acceleration is adjusted to diminish the velocity:
\begin{equation}\label{eq:evasion}
a_{y}=min( max(-v_y/\delta t, -a_{y,m}), a_{y,m}),
\end{equation}
where $a_{y,m}$ is the absolute value of the maximal lateral acceleration.


Next we will analyze the safe evasion trajectory. The optimal lateral motion is to return back to the original lane as soon as possible. We define time $t=0$ when the ego vehicle just starts taking the evasion trajectory. The centers of the original and the target lane are $y=0$ and $y=w_l$, respectively. The width of a vehicle is denoted as $w_v$. The ego vehicle is completely in the original lane when $p_y \leq \frac{w_l-w_v}{2}$. 
The fastest lateral motion is that the ego vehicle has lateral acceleration $a_y=-a_{y,m}$ when $t \in [0, t_1]$ and then $a_y=a_{y,m}$ when $t \in [t_1, t_{y,f}]$, 
and finally it reaches the position $p_y=\frac{w_l-w_v}{2}$ with $v_y=0$:
\begin{equation}\label{eq:lateral time_0}
\begin{cases}
p_{y,t_0}+v_{y,t_0} t_1 - \frac{a_{y,m} t_1^2}{2}+(v_{y,t_0}-a_{y,m} t_1) (t_{y,f}-t_1) 
\\ \quad \quad +
\frac{a_{y,m}(t_{y,f}-t_1)^2}{2}=\frac{w_l-w_v}{2}
\\
v_{y,t_0}-a_{y,m} t_1 + a_{y,m}(t_{y,f}-t_1)=0
\end{cases}
\end{equation}
Here $p_{y,t_0}$ and $v_{y,t_0}$ are the lateral position and velocity of the ego vehicle when $t=0$. 

We assume that the ego vehicle $E$, leading vehicle $L$ and following vehicle $F$ all have the same maximal longitudinal acceleration $a_{x,a}=a_{x,l,a}=a_{x,f,a}$ and braking deceleration $a_{x,d}=a_{x,l,d}=a_{x,f,d}$. Let $p_{x,t_0}$ and $v_{x,t_0}$ represent the longitudinal position and velocity of the ego vehicle when $t=0$, respectively. For simplicity of notation, we omit the subscript of $t_0$, and use $p_{x,l}$, $v_{x,l}$, $p_{x,f}$ and $v_{x,f}$ to denote the longitudinal position and velocity of the leading and following vehicles when $t=0$, respectively. Next we will temporarily neglect the following vehicle $F$, and analyze the longitudinal motion when the leading vehicle $L$ decelerates with $a_{x,l,d}$ abruptly. 



If $v_{x,l} \geq v_{x,t_0}$, then $\frac{v_{x,l}^2}{2a_{x,l,d}}\geq \frac{v_{x,t_0}^2}{2a_{x,d}}$, and the ego vehicle can prevent collisions with the leading vehicle if it decelerates with $a_{x,d}$. If $v_{x,l}<v_{x,t_0}$, the ego vehicle takes at least $t_{x,f}=\frac{v_{x,t_0}}{a_{x,d}}$ to fully stop. If $t_{x,f} \geq t_{y,f}$, ego vehicle only needs to keep enough headway when $t \in [0, t_{y,f}]$ because it is already not in the target lane when $t \in [t_{y,f}, t_{x,f}]$. We define $C_1$ to reflect the minimum headway\footnote{It is indeed a constant minus the minimum headway.}:

\begin{equation}\label{eq:leading vehicle brakes_1}
C_1=
\begin{cases}
p_{x,t_0}-p_{x,l}+v_{x,t_0}t_{y,f}-\frac{a_{x,d}t_{y,f}^2}{2}-\frac{v_{x,l}^2}{2a_{x,l,d}}+p_{m}
\\
\qquad \qquad \text{if } \frac{v_{x,l}}{a_{x,l,d}}<t_{y,f}
\\
p_{x,t_0}-p_{x,l}+(v_{x,t_0}-v_{x,l})t_{y,f}-\frac{(a_{x,d}-a_{x,l,d})t_{y,f}^2}{2}+p_{m}
\\
\qquad \qquad \text{otherwise}
\end{cases}
\end{equation}
Here $p_{m}$ is the minimum gap between vehicles to avoid collisions. We need $C_1<0$ to ensure safety. If $t_{x,f} < t_{y,f}$, ego vehicle needs to keep enough headway when $t \in [0, t_{x,f}]$. Similarly, we define $C_2$ and need $C_2<0$ to ensure safety:
\begin{equation}\label{eq:leading vehicle brakes_2}
C_2=p_{x,t_0}-p_{x,l}+\frac{v_{x,t_0}^2}{2a_{x,d}}-\frac{v_{x,l}^2}{2a_{x,l,d}}+p_{m}
\end{equation}




Next we assume that at the same time, the following vehicle $F$ accelerates to prevent the ego vehicle from cutting in. In such case, the ego vehicle can even accelerate and get closer to the leading vehicle, thus acquiring more time for lateral evasion before the following vehicle catches up. 
It is indeed the fastest longitudinal motion to prevent collision with the following vehicle by first accelerating with $a_{x,a}$, and then decelerating with $a_{x,d}$. 


We assume that the ego vehicle accelerates with $a_x=a_{x,a}$ when $t\in[0, t_2]$ and then decelerates with $a_x=-a_{x,d}$ until it stops when $t \in [t_2, t_{y,f}]$. By letting the minimum distance between ego vehicle and leading vehicle be exactly $p_m$ to remain safe, $t_2$ can be represented as a function of $C_1$ and $C_2$.

\begin{table*}[]
\centering
\caption{Safety and performance evaluation for our proposed framework. From top to bottom, experimental settings correspond to more challenging lane changing scenarios. Our `SafIn NN' planner results in zero collision rate in all simulations.
}
\label{table: statistical evaluation}
\begin{tabular}{|C{2cm}|c|C{2cm}|C{2cm}|C{2cm}|C{2cm}|}
\hline
                         experimental settings             & methods        & lane changing time & final lateral position & success rate & collision rate \\ \hline
\multirow{3}{3cm}{$-6 \leq a_{x,l}\leq 4$, $7 \leq \delta p \leq 37$} & MPC            &    1.90 s      &     \textbf{3.44 m}       &    \textbf{92.61\%}      &     7.39\%           \\ \cline{2-6} 
                                                                    & only NN        &    \textbf{1.70 s}        &     3.25 m       &   89.59\%       &   10.41\%             \\ \cline{2-6} 
                                                                    & SafIn NN &    1.90 s        &     2.73 m       &   80.31\%       &   \textbf{0\%}             \\ \hline
\multirow{3}{3cm}{$-6 \leq a_{x,l}\leq 0$, $7 \leq \delta p \leq 37$} & MPC            &     1.90 s     &     \textbf{3.46 m}       &    \textbf{87.46\%}      &   12.54\%             \\ \cline{2-6} 
                                                                    & only NN        &     \textbf{1.68 s}       &     3.30 m       &     82.37\%     &   17.63\%             \\ \cline{2-6} 
                                                                    & SafIn NN &     2.08 s       &     2.44 m       &     67.89\%     &   \textbf{0\%}             \\ \hline
\multirow{3}{3cm}{$-6 \leq a_{x,l}\leq 4$, $7 \leq \delta p \leq 17$} & MPC            &    1.90 s      &    \textbf{3.44 m}        &     83.06\%     &  16.94\%              \\ \cline{2-6} 
                                                                    & only NN        &    \textbf{1.73 s}        &    3.24 m        &     \textbf{84.53\%}     &   15.47\%             \\ \cline{2-6} 
                                                                    & SafIn NN &    1.97 s        &    2.23 m        &     61.51\%     &   \textbf{0\%}             \\ \hline
\multirow{3}{3cm}{$-6 \leq a_{x,l}\leq 0$, $7 \leq \delta p \leq 17$} & MPC            &    1.90 s      &    \textbf{3.46 m}        &    71.82\%      &   28.18\%             \\ \cline{2-6} 
                                                                    & only NN        &    \textbf{1.71 s}        &    3.29 m        &    \textbf{74.32\%}      &  25.68\%              \\ \cline{2-6} 
                                                                    & SafIn NN &    2.34 s        &    1.66 m        &    38.76\%      &  \textbf{0\%}              \\ \hline
\end{tabular}
\end{table*}



To prevent collisions with the following vehicle accelerating with $a_{x,f,a}$, we have
\begin{equation}\label{eq:following vehicle accelerates_4}
\begin{cases}
p_{x,t_0}+v_{x,t_0}t_2+\frac{a_{x,a}t_2^2}{2}+\frac{(v_{x,t_0}+a_{x,a}t_2)^2}{2a_{x,d}}
\\
-p_{x,f}-v_{x,f}t_{y,f}-\frac{a_{x,f,a}t_{y,f}^2}{2}-p_{m} > 0
\\
\qquad \qquad \text{if } t_2+\frac{v_{x,t_0}+a_{x,a}t_2}{a_{x,d}}<t_{y,f}
\\
p_{x,t_0}+v_{x,t_0}t_2+\frac{a_{x,a}t_2^2}{2}+(v_{x,t_0}+a_{x,a}t_2)(t_{y,f}-t_2)
\\
-\frac{a_{x,d}(t_{y,f}-t_2)^2}{2}-p_{x,f}-v_{x,f}t_{y,f}-\frac{a_{x,f,a}t_{y,f}^2}{2}-p_{m} > 0
\\
\qquad \qquad \text{otherwise}
\end{cases}
\end{equation}

Similar to Eq.~\ref{eq:following vehicle accelerates_4}, we can derive the inequality constraint for ensuring system safety when the following vehicle is cautious, willing to decelerate and create gap for ego vehicle.
In summary, if the constraints can be satisfied, the ego vehicle is verified to have safe evasion trajectory after taking planned trajectory; otherwise, it has to adjust the driving behavior to prevent possible collisions.

\section{Experimental Results}\label{sec:experiments}
In this section, we first demonstrate the effectiveness of our proposed framework via simulations on synthetic and real-world examples. We compare our approach (denoted as `SafIn NN') with a neural network-based planner without safety consideration or interactive planning (denoted as `only NN') and an MPC-based planner from~\cite{2020arXiv200108620L}. 
We then further evaluate the performance of our aggressiveness assessment module for the following vehicle in the target lane. 


\subsection{Evaluation with Synthetic Examples}\label{sec:evaluation simulations}

We first evaluate the performance of our proposed framework through extensive simulations on synthetic examples, and the results are shown in Table~\ref{table: statistical evaluation}. We have four classes with different experimental settings, which indicate the ranges that $a_{x,l}$ and $\delta p$ will uniformly sample from. $\delta p$ is the initial longitudinal distance between leading vehicle and ego vehicle. Thus $-6 \leq a_{x,l}\leq 4$ and $7 \leq \delta p \leq 37$ correspond to easier lane changing scenarios, while $-6 \leq a_{x,l}\leq 0$ and $7 \leq \delta p \leq 17$ correspond to more congested and dangerous scenarios. For each class, we conduct 200,000 simulations with randomly generated relative positions, velocities and IDM parameters. The following vehicle has 50\% probability of being aggressive, and another 50\% probability of being cautious. 





\begin{figure*}[tbp]
\centering\includegraphics[scale=0.3]{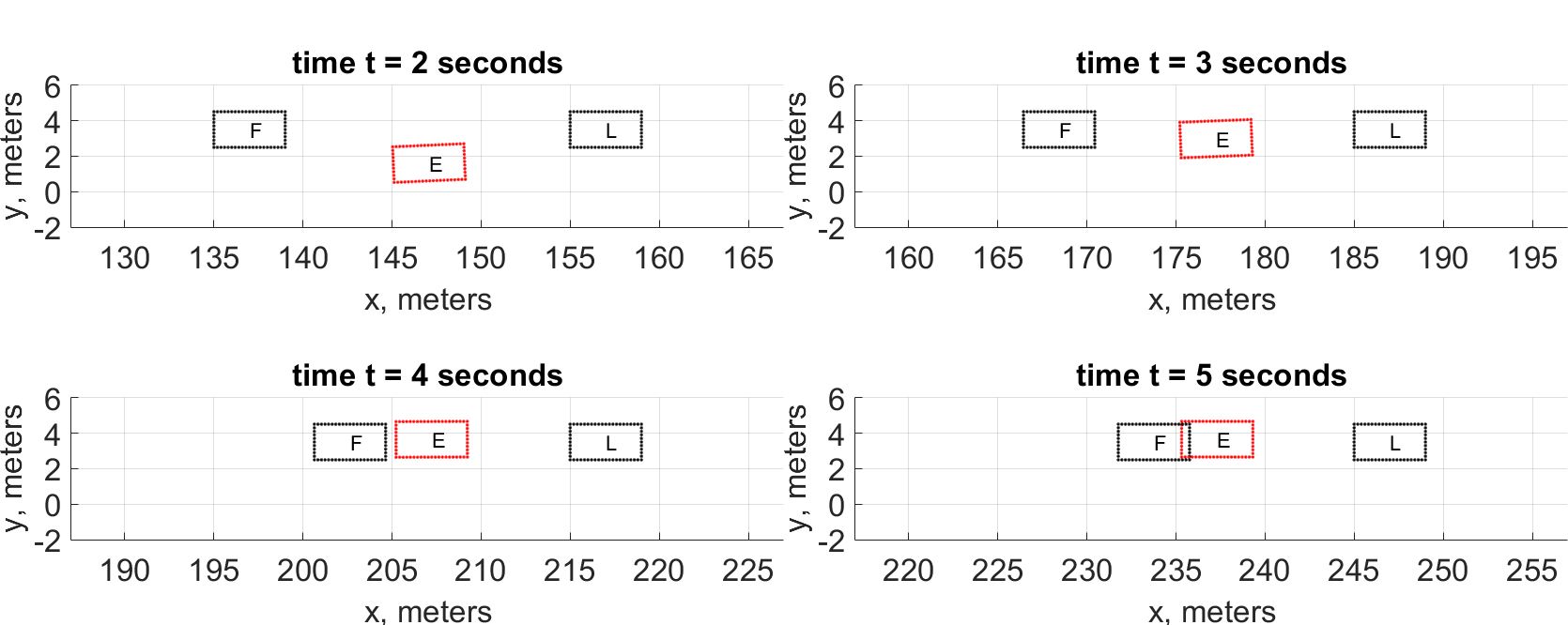}
\caption{Illustrating Example. The x and y axes show the longitudinal and lateral positions of vehicles. Four subplots show the positions at different times. Red rectangle represents the ego vehicle, and black rectangles are surrounding vehicles. It corresponds to the scenario that initial velocity is $30$ m/s for all vehicles, and the distance between the leading vehicle and the following vehicle is $20$ m. The following vehicle accelerates at $t=2$ seconds to prevent the ego vehicle from cutting in. The ego vehicle is controlled by the `only NN' planner.}
\label{fig:acc_no_adj}
\end{figure*}


\begin{figure}[tbp]
\centering\includegraphics[scale=0.45]{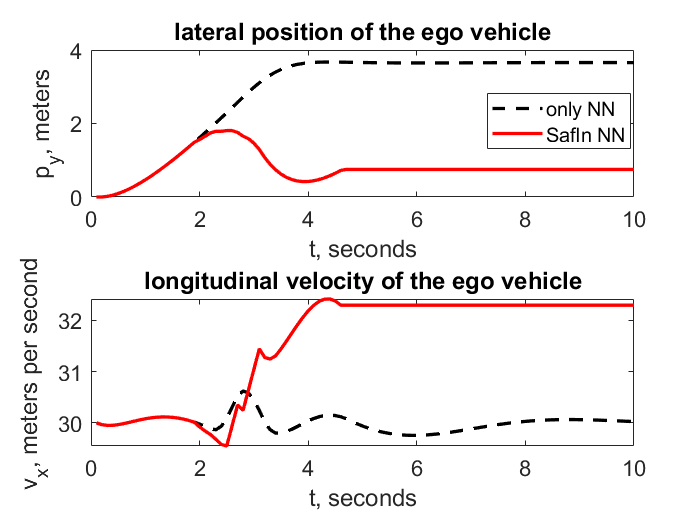}
\caption{Lateral position and longitudinal velocity of the ego vehicle in the same scenario as in Fig.~\ref{fig:acc_no_adj}.}
\label{fig:acc_ego_status}
\end{figure}






For every round simulation with a horizon of 10 seconds, the ego vehicle attempts to change lanes until it collides with other vehicles. It is successful if the ego vehicle finally crosses the two lanes within the simulation horizon without any collision. For all successful lane changing simulations, we compute the average time that it takes to cross the lanes. For all safe simulations, we compute the average of final lateral positions. The lateral position $y=3.5$ meters represents the center of target lane, and $y=1.75$ meters is the border of two lanes.

Table~\ref{table: statistical evaluation} shows that: (1) Our approach `SafIn NN' results in \textbf{zero collision rate in all simulations regardless whether the following vehicle is aggressive or not}, while `MPC' and `only NN' both lead to significant collision rate, especially in more challenging scenarios. (2) Under `MPC' and `only NN', the ego vehicle has a higher lane changing success rate, less lane changing time and larger final lateral position. In easier scenarios, these advantages are relatively small. When it becomes more congested and challenging, risky behaviors are restricted under our `SafIn NN' planner while lead to higher collision rate for 'MPC' and 'only NN'.


We further demonstrate the strength of our `SafIn NN' planner with a concrete example. The initial longitudinal velocities of all three vehicles are set to 30 meters per second. The distance between leading and following vehicle is 20 meters. Then we let the following vehicle start accelerating with $a_{x,f}=4$ meters per second squared until $p_{x,l}-p_{x_f} \leq 17$ meters at $t=2$ seconds. Then it adjusts velocity and attempts to maintain the distance to the leading vehicle. Fig.~\ref{fig:acc_no_adj} presents the relative position changes of vehicles when the ego vehicle is controlled by `only NN' planner. The ego vehicle completes lane changing at $t=3$ seconds, but collides with the following vehicle at $t=5$ seconds. Fig.~\ref{fig:acc_ego_status} shows the lateral position and longitudinal velocity of the ego vehicle under `only NN' and `SafIn NN' planners, respectively. Our `SafIn NN' planner prevents the collision proactively. The ego vehicle aborts changing lanes at around $t=3$ seconds, and then hesitates around $y=1$ meters and looks for the next chance to change lanes. %

\vspace{-12pt}

\subsection{Evaluation with Real-world Challenging Dataset}

We further evaluate our approach in challenging scenarios with real-world dataset collected by Pony.ai, an autonomous vehicle company. The dataset provides road geometry and motion information of surrounding traffic participants in congested scenarios with industry-level accuracy. Under our designed planner, the ego vehicle is controlled to change lanes. \textbf{In all tested 48 real-world challenging scenarios, the ego vehicle can always remain safe under our planner} during the lane change process, despite that our planner is never trained or optimized with the dataset. Under `only NN' planner, there are 12 scenarios in which the ego vehicle collides with other vehicles.

\begin{figure}[tbp]
\centering\includegraphics[scale=0.45]{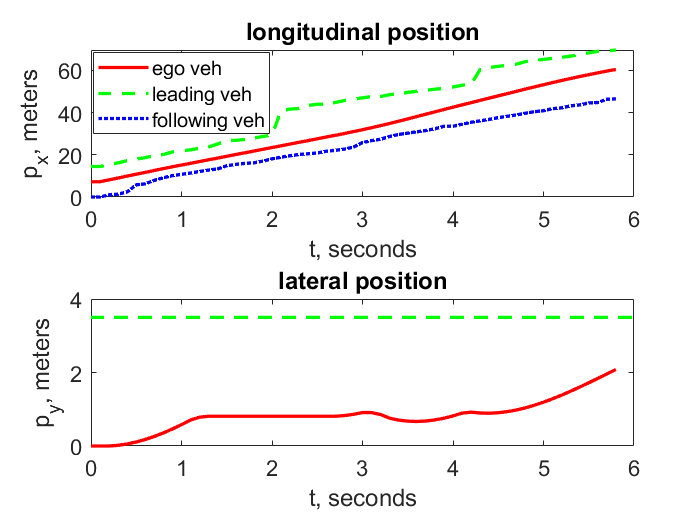}
\caption{Longitudinal and lateral position of the ego vehicle, leading vehicle and following vehicle. It shows an example of challenging scenario in dataset collected by Pony.ai and the ego vehicle is controlled by the `SafIn NN' planner.}
\label{fig:example_18}
\end{figure}

Fig.~\ref{fig:example_18} shows a concrete example of the challenging scenario that the gap between the leading vehicle and the following vehicle is decreasing initially. The ego vehicle under our designed planner attempts to change lane at the beginning and then hesitates around $p_y=0.8$ meters from $t=1.1$ seconds, and finally continues changing lane from $t=4.5$ seconds. While under the `only NN' planner, the ego vehicle will collide with the following vehicle at $t=1.6$ seconds in this challenging scenario.

\subsection{Evaluation of Aggressiveness Assessment}
\begin{table}[]
\centering
\caption{Performance of aggressiveness assessment.}
\label{table: agg results}
\scalebox{0.9}{
\begin{tabular}{cccccc}
\hline
\multicolumn{1}{c|}{}               & \multicolumn{1}{c|}{$a_{th}=0$} & \multicolumn{1}{c|}{$a_{th}=0.15$} & \multicolumn{1}{c|}{$a_{th}=0.25$} & \multicolumn{1}{c|}{$a_{th}=0.5$} & $a_{th}=1$ \\ \hline
easy                                &                                 &                                    &                                    &                                   &              \\ \hline
\multicolumn{1}{c|}{uncertain rate} & \multicolumn{1}{c|}{0\%}           & \multicolumn{1}{c|}{2.28\%}              & \multicolumn{1}{c|}{4.05\%}              & \multicolumn{1}{c|}{9.16\%}             & \multicolumn{1}{c}{19.3\%}\\
\multicolumn{1}{c|}{error rate}       & \multicolumn{1}{c|}{4.61\%}           & \multicolumn{1}{c|}{3.6\%}              & \multicolumn{1}{c|}{3.05\%}              & \multicolumn{1}{c|}{2.01\%}             & \multicolumn{1}{c}{0.91\%}\\ \hline
medium                              &                                 &                                    &                                    &                                   &              \\ \hline
\multicolumn{1}{c|}{uncertain rate} & \multicolumn{1}{c|}{0\%}           & \multicolumn{1}{c|}{18.38\%}              & \multicolumn{1}{c|}{31.33\%}              & \multicolumn{1}{c|}{56.32\%}             & \multicolumn{1}{c}{79.65\%}\\
\multicolumn{1}{c|}{error rate}       & \multicolumn{1}{c|}{36.73\%}           & \multicolumn{1}{c|}{28.82\%}              & \multicolumn{1}{c|}{24.53\%}              & \multicolumn{1}{c|}{15.87\%}             & \multicolumn{1}{c}{7.16\%}\\ \hline
hard                                &                                 &                                    &                                    &                                   &              \\ \hline
\multicolumn{1}{c|}{uncertain rate} & \multicolumn{1}{c|}{0\%}           & \multicolumn{1}{c|}{65.26\%}              & \multicolumn{1}{c|}{74.16\%}              & \multicolumn{1}{c|}{85.57\%}             & \multicolumn{1}{c}{94.39\%}\\
\multicolumn{1}{c|}{error rate}       & \multicolumn{1}{c|}{49.76\%}           & \multicolumn{1}{c|}{17.18\%}              & \multicolumn{1}{c|}{12.76\%}              & \multicolumn{1}{c|}{7.1\%}             & \multicolumn{1}{c}{2.74\%}\\ \hline
\end{tabular}
}
\end{table}

We conduct experiments to evaluate the performance of our aggressive assessment module, and Table~\ref{table: agg results} shows the results. We classify all simulation entries in the dataset into three classes based on the difference of accelerations under different behaviors, $\delta a_{x,f}^*= |a_{x,f,1}^*-a_{x,f,2}^*|$. It is classified as easy, medium or hard if $\delta a_{x,f}^*>0.5$, $0.25 < \delta a_{x,f}^* \leq 0.5$, or $\delta a_{x,f}^* \leq 0.25$, respectively. We conduct sensitivity analysis over the threshold $a_{th}$. It shows that with larger $a_{th}$, the uncertain rate is higher and the error rate is lower for all three different difficulty levels. It meets our expectation because a larger $a_{th}$ results in a more robust and conservative predictor, which is prone to be uncertain when it is less confident. It also presents that for easy cases, the performance can be considerably greater because a larger $\delta a_{x,f}^*$ means that it is more distinguishable.

For simulations conducted in Section~\ref{sec:evaluation simulations}, we use $a_{th}=0.5$. Although it has a positive error rate, our overall approach is quite robust and does not result in collisions in all experiments. We think that there are two reasons: (1) aggressiveness assessment is conducted every 0.1 seconds along with other modules, and thus occasional mis-prediction is highly likely to be corrected later; and (2) it is more challenging to make correct assessment when the following vehicle is far away from the ego vehicle. However, in those cases, incorrect assessment is less critical because of the large gap between vehicles. 

\subsection{Discussion on MPC and Neural Network-based Planners}
In this work, the neural network planners are learned from the synthesized data of the system under MPC, and thus `only NN' has similar performance as MPC -- albeit in more challenging scenarios, `only NN' shows slight advantages in both success rate and collision rate. Moreover, MPC with the safety-driven behavior adjustment module also provide similar performance as our `SafIn NN' planner. However, note that our safety-driven interactive planning framework can be incorporated with any state-of-the-art neural network-based planners to improve safety. We believe that with more high-quality training data, `SafIn NN' can also significantly improve its performance in success rate and may perform better than MPC-based planners in all metrics, especially when system dynamics and interactions are hard to model (in this work the MPC is assumed to have perfect system model).

\section{Conclusion}\label{sec:conclusion}
In this work, we present a novel safety-driven interactive planning framework for neural network-based lane changing. The framework includes a safety-driven behavior adjustment module for safety assurance and an aggressiveness assessment module for avoiding over-conservative planning. Extensive experiments on synthetic examples and real-world challenging scenarios demonstrate the effectiveness of our approach in improving system safety. In future work, we will continue improving the performance (success rate) of our approach while ensuring safety. We plan to start with improving the neural network-based planners by conducting training on higher quality data and with more advanced methods.





\bibliographystyle{ACM-Reference-Format}
\bibliography{Refs}


\end{document}